\documentclass[conference]{IEEEtran}
\IEEEoverridecommandlockouts
\usepackage{cite}
\usepackage{amsmath,amssymb,amsfonts}
\usepackage{algorithmic}
\usepackage{graphicx}
\usepackage{textcomp}
\usepackage{xcolor}
\usepackage{multirow}
\usepackage{jabbrv}
\usepackage[super]{nth}
\def\BibTeX{{\rm B\kern-.05em{\sc i\kern-.025em b}\kern-.08em
    T\kern-.1667em\lower.7ex\hbox{E}\kern-.125emX}}

\makeatletter 
\newcommand{\linebreakand}{%
  \end{@IEEEauthorhalign}
  \hfill\mbox{}\par
  \mbox{}\hfill\begin{@IEEEauthorhalign}
}
\makeatother 

\begin{document}

\title{Brain-Driven Representation Learning Based on Diffusion Model
\footnote{{\thanks{This work was supported by Institute for Information \& Communications Technology Planning \& Evaluation (IITP) grant funded by the Korea government (MSIT) (No.2021-0-02068, Artificial Intelligence Innovation Hub; No. 2019-0-00079, Artificial Intelligence Graduate School Program(Korea University)).}
}}
}

\author{

\IEEEauthorblockN{Soowon Kim}
\IEEEauthorblockA{\textit{Dept. of Artificial Intelligence} \\
\textit{Korea University} \\
Seoul, Republic of Korea \\
soowon\_kim@korea.ac.kr}  \\

\and

\IEEEauthorblockN{Seo-Hyun Lee}
\IEEEauthorblockA{\textit{Dept. of Brain and Cognitive Engineering} \\
\textit{Korea University} \\
Seoul, Republic of Korea \\
seohyunlee@korea.ac.kr} \\

\and

\IEEEauthorblockN{Young-Eun Lee}
\IEEEauthorblockA{\textit{Dept. of Brain and Cognitive Engineering} \\
\textit{Korea University} \\
Seoul, Republic of Korea \\
ye\_lee@korea.ac.kr} \\

\linebreakand 

\IEEEauthorblockN{Ji-Won Lee}
\IEEEauthorblockA{\textit{Dept. of Artificial Intelligence} \\
\textit{Korea University} \\
Seoul, Republic of Korea \\
jiwon\_lee@korea.ac.kr} \\

\and

\IEEEauthorblockN{Ji-Ha Park}
\IEEEauthorblockA{\textit{Dept. of Artificial Intelligence} \\
\textit{Korea University} \\
Seoul, Republic of Korea \\
jiha\_park@korea.ac.kr} \\

\and 

\IEEEauthorblockN{Seong-Whan Lee}
\IEEEauthorblockA{\textit{Dept. of Artificial Intelligence} \\
\textit{Korea University} \\
Seoul, Republic of Korea \\
sw.lee@korea.ac.kr}
}










\maketitle

\begin{abstract}

Interpreting EEG signals linked to spoken language presents a complex challenge, given the data's intricate temporal and spatial attributes, as well as the various noise factors. Denoising diffusion probabilistic models (DDPMs), which have recently gained prominence in diverse areas for their capabilities in representation learning, are explored in our research as a means to address this issue. Using DDPMs in conjunction with a conditional autoencoder, our new approach considerably outperforms traditional machine learning algorithms and established baseline models in accuracy. Our results highlight the potential of DDPMs as a sophisticated computational method for the analysis of speech-related EEG signals. This could lead to significant advances in brain-computer interfaces tailored for spoken communication.

\end{abstract}

\begin{small}
\textbf{\textit{Keywords--brain-computer interface, electroencephalogram, imagined speech, diffusion model;}}\\
\end{small}

\section{INTRODUCTION}

Speech serves as an essential means of human communication, allowing us to express intricate thoughts and ideas through audible patterns. Speech capacity is deeply embedded in our social and cultural fabric, facilitating everything from relationship building to information sharing. Despite its importance, some people, such as those suffering from locked-in syndrome, are unable to engage in verbal communication due to physical limitations \cite{thung2018conversion}. Therefore, innovative approaches to restore or replace speech capabilities are a vital research frontier. In line with this, our work focuses on the interpretation of brain signals as a means for facilitating non-vocal communication.

Electroencephalography (EEG) offers a non-invasive avenue for capturing the brain's electrical activities. Acquired through scalp-placed electrodes, EEG signals are instrumental in various applications, ranging from neuroscience to clinical diagnostics \cite{kim2015abstract}. The decoding of these EEG signals into useful data, such as speech-related activities or focus levels, is of growing interest.

Deciphering EEG data related to spoken language is notably intricate. The task involves interpreting complex and highly variable neural activities related to the articulation and perception of speech. Additionally, these EEG signals often contain noise and artifacts, further complicating accurate decoding. In light of these challenges, ongoing research aims to establish robust and effective methods for EEG signal interpretation, which have broad applications including speech restoration and human-machine interactions.

Denoising diffusion probabilistic models (DDPMs) have emerged as a potent tool for identifying nuanced patterns within complicated, high-dimensional datasets. Through a process of adding Gaussian noise over a series of steps, DDPMs corrupt an original signal and then attempt to reconstruct it. These models have been particularly successful in dealing with time series data, including audio and video streams.

Decoding EEG signals using deep learning approaches is a challenging problem due to various factors, including the scarcity of data, a poor signal-to-noise ratio, and high inter- and intra-individual variability \cite{tayeb2019validating}. Despite these challenges, several studies have explored different EEG decoding techniques for various applications, including speech decoding \cite{lee2019possible}.

On the basis of existing research, various methods for the decoding of EEG signals have been explored. Schirrmeister et al. utilized DeepConvNets \cite{schirrmeister2017deep} to achieve end-to-end learning in human EEG signals, using machine learning techniques such as batch normalization and exponential linear units. They achieved performance on par with traditional filter bank common spatial pattern algorithms. Lawhern et al. \cite{kim2019subject,mane2021fbcnet} introduced EEGNet, a specialized CNN architecture for EEG classification, using depthwise and separable convolutions to better capture specific EEG features. This model has been successfully tested on multiple BCI paradigms.

Furthermore, Lee et al. \cite{ho2020denoising, lee2019towards} conducted an in-depth study of the nuances that affect decoding performance in two key BCI paradigms: imagined speech and visual imagery. The study used EEG signals filtered across multiple frequency ranges and identified relevant cortical regions \cite{suk2014predicting}, resulting in high precision and multiclass scalability for both paradigms.

In parallel, diffusion-based approaches for time series data have gained substantial traction. One such method, proposed by Alcaraz et al., uses a structured state-space model with an integrated diffusion process for time-series data imputation and forecasting, showing superior performance to existing methods. Jeong et al. offered a novel application of diffusion models to improve synthetic speech quality in Text-to-Speech (TTS) systems, which also demonstrated effectiveness over current methods.

Our study builds on this background to introduce a new strategy for interpreting EEG signals linked to spoken language using DDPM and a conditional autoencoder (CAE). The CAE facilitates the retention of valuable features that may otherwise be compromised during the DDPM's forward process. Additionally, we incorporate a jointly trained classifier to enhance decoding performance. To the best of our knowledge, this is the first effort to apply diffusion models to interpret speech-related EEG signals.

Our study extends this body of work by introducing a novel approach that combines DDPMs and a conditional autoencoder to decode EEG signals related to spoken language. This method aims to capture the intricate neural patterns and relationships inherent in speech processes, and, in doing so, advances the field of EEG decoding with potential applications in speech rehabilitation and brain-computer interfaces.

\begin{figure}[t]
  \centering
  \includegraphics[width=\linewidth,height=\textheight,keepaspectratio]{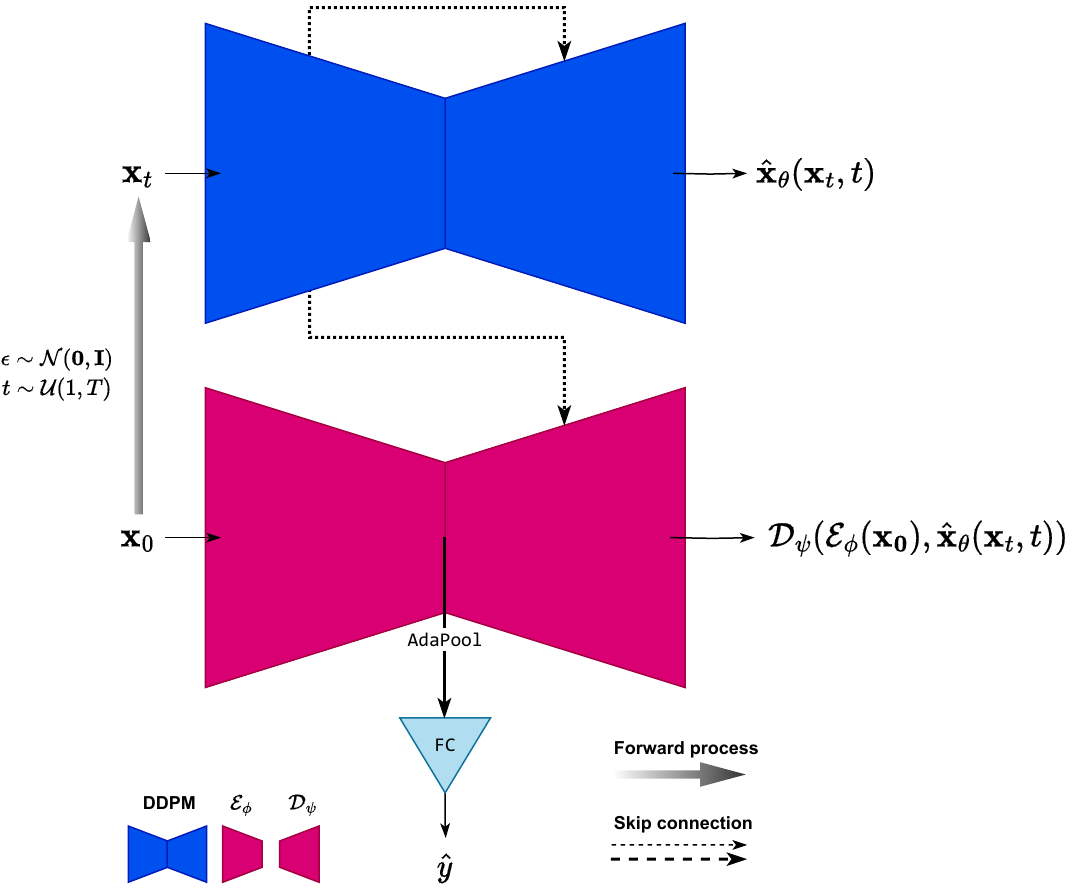}
  \caption{A flowchart of Diff-E for EEG signal decoding. Initially, DDPM processes noisy data to approximate the original signal, then a CAE refines this output by correcting the discrepancies. Subsequently, the classifier utilizes the encoder's output for downstream classification tasks, enhancing overall performance.}
  \label{fig:method}
\end{figure}

\section{MATERIALS AND METHODS}
\subsection{Denoising Diffusion Models}
DDPMs are a type of machine learning model that can learn complex probability distributions over data. The "forward process" in DDPMs is determined by a fixed Markov chain that progressively adds Gaussian noise to the data. The forward process begins with a probability distribution, denoted $q(\mathbf{x}_0)$, which represents the uncorrupted original data. This distribution is then iteratively transformed using a sequence of Markov diffusion kernels, $q(\mathbf{x}_t \mathbf{x}_{t-1})$, which are Gaussian with a fixed variance schedule $\{\beta_t\}^T_{t=1}$. This process can be expressed as follows:

\begin{equation}
    q(\mathbf{x}_t|\mathbf{x}_{t-1})=\mathcal{N}(\mathbf{x}_t;\sqrt{1-\beta_t}\mathbf{x}_{t-1},\beta_t\mathbf{I}),
\end{equation}
\hspace{0.5cm}
\begin{equation}
    q(\mathbf{x}_{1:T}|\mathbf{x}_0)=\prod_{t=1}^{T}q(\mathbf{x}_t|\mathbf{x}_{t-1}).
\end{equation}

The original data can be corrupted by a diffusion process with Gaussian noise at any stage, $t$, where $\alpha_t$ is represented as $1-\beta_t$ and $\Bar{\alpha}_t$ is the product of all $\alpha_s$ from $s=1$ to $t$.

\begin{equation}
    q(\mathbf{x}_t|\mathbf{x}_0)=\mathcal{N}(\mathbf{x}_t;\sqrt{\Bar{\alpha}_t}\mathbf{x}_0, (1-\Bar{\alpha}_t) \mathbf{I}).
\end{equation}

Ho et al. \cite{ho2020denoising} proposed a technique to train a model that takes a noisy sample $\mathbf{x}_t$ and predicts the noise it contains by training a network $\epsilon\theta(\mathbf{x}_t,t)$. On the contrary, our research trains Diff-E to forecast the original unchanged signal, $\mathbf{x}_0$, rather than predicting the injected noise.

\begin{equation}
\mathcal{L}_{\text{DDPM}}(\theta) = ||\mathbf{x}_0 - \hat{\mathbf{x}}_\theta(\mathbf{x}_t,t)||.
\label{eq:DDPM}
\end{equation}

We randomly select a timestep $t$ from a uniform distribution, $\mathcal{U}(1, T)$, and use $\theta$ as the parameters of the DDPM. The objective of the model is to denoise the noisy input and generate an output that is close to the original signal. We have employed a time-conditional UNet architecture \cite{ronneberger2015u}, similar to the one used in \cite{ho2020denoising}, with modifications to make it suitable for EEG data. The DDPM's prediction is denoted as $\hat{\mathbf{x}}_\theta(\mathbf{x}_t,t)$.

\subsection{Conditional Autoencoder}
The DDPM's forward pass leads to information loss, which the CAE attempts to make up for by recognizing and correcting these errors. This allows the CAE to generate more precise representations of the original EEG signals. To this end, we use the following objective function for the CAE:
\begin{equation}
    \mathcal{L}_{\text{CAE}}(\psi, \phi) = ||\mathcal{L}_{\text{DDPM}}(\theta) -\mathcal{D}_\psi(\mathcal{E}_\phi(\mathbf{x}_0),\hat{\mathbf{x}}_\theta(\mathbf{x}_t, t))||.
\label{eq:CAE}
\end{equation}

The CAE includes an encoder and decoder, denoted as $\mathcal{E}_\phi$ and $\mathcal{D}_\psi$, respectively. $\mathcal{D}_\psi$ is connected to the DDPM layers instead of the output of $\mathcal{E}_\phi$. This allows $\mathcal{D}_\psi$ to be implicitly conditioned on the corruption stage of the DDPM, as illustrated in Fig. \ref{fig:method} with \textit{dashed arrows}. Additionally, to improve the reconstruction of $\mathcal{L}_{\text{DDPM}}$, the original signal, $\mathbf{x}_0$, and the output of the DDPM, $\hat{\mathbf{x}}_\theta(\mathbf{x}_t,t)$, are connected to the last layer of $\mathcal{D}_\psi$ as shown in Fig. \ref{fig:method} with \textit{thin dashed arrows}.

\subsection{Classifier}
After $\mathcal{E}_\phi$ has processed the data, the output is condensed into a single-dimensional representation, $\mathbf{z}$, using an adaptive average pooling layer. This creates a latent vector which is then fed into the linear classifier $\mathcal{C}_\rho$. The classifier is trained jointly with the CAE to differentiate the representations of each class and classify them. The dimension of $\mathbf{z}$ is fixed at $256$ for the duration of the experiment. To include the classification loss in the CAE's objective function, we modified it to become the overall Diff-E objective.

\begin{equation}
    \begin{split}
        \mathcal{L}_{\text{Diff-E}}(\psi, \phi, \rho) &= ||\mathcal{L}_{\text{DDPM}}(\theta) -\mathcal{D}_\psi(\mathcal{E}_\phi(\mathbf{x}_0),\hat{\mathbf{x}}_\theta(\mathbf{x}_t, t))||\\
        &+\alpha||\hat{y}-y||_2.
    \end{split}
\label{eq:Diff-E}
\end{equation}

The predicted label of the input signal is calculated as $\hat{y}=\mathcal{C}_\rho(\mathbf{z})$, where $\rho$ is an adjustable parameter for $\mathcal{C}_\rho$ and $y$ is the true label. The hyperparameter $\alpha$ is used to control the relative importance of the reconstruction loss and the classification loss, with a value of $0.1$ chosen for the experiment. During inference, only $\mathcal{E}_\phi$ and $\mathcal{C}_\rho$ are used to classify the signals, with the predicted label obtained as $\hat{y}=\mathcal{E}_\phi(\mathcal{C}_\rho(\mathbf{x}_0))$. To evaluate the effectiveness of Diff-E, it is compared to other methods that have been applied to decoding EEG signals in various paradigms, such as motor imagery and event-related potentials \cite{bang2021spatio,Lee2019comparative}. This comparison is conducted to assess the performance of Diff-E and to determine its suitability for imagined speech EEG signal decoding applications.

\subsection{Model Implementation Details}

In our study, the DDPM and CAE frameworks are constructed with layers that sequentially execute convolution, normalization, and activation functions. The encoder employs adaptive pooling to produce a compact feature vector, \(\mathbf{z}\). The total number of trainable parameters for DDPM and CAE is roughly $3\mathrm{e}{+5}$ and the classifier has $4\mathrm{e}{+5}$ parameters. Optimization is conducted using RMSProp and a cyclic learning rate that starts at $9\mathrm{e}{-5}$ and caps at $1.5\mathrm{e}{-3}$. The training extended over 500 epochs, with L1 loss for DDPM and CAE, and mean squared error for the classifier's one-hot encoded classification tasks. For model evaluation, 20 \% of the data was reserved for testing, with a consistent random seed ensuring reproducibility.

\subsection{Dataset}
\subsubsection{Data Description}
This study used data from a previous study by Lee et al. \cite{krepki2007berlin, lee2020neural}. Participants were 22 healthy individuals, 15 of whom were male, with an average age of 24.68 ± 2.15. None of them had a history of neurological disease or language disorders and had no hearing or visual impairments. Furthermore, they did not take drugs for 12 hours prior to the session. All of them had received high-quality English education for more than 15 years. The overt speech task involved instructing the 22 subjects to imagine saying 12 different words or sentences, such as ``ambulance," ``clock," ``hello," ``help me," ``light", ``pain," ``stop," ``thank you," ``toilet," ``TV," ``water" and ``yes," as well as a resting state, resulting in a total of 13 classes. The researchers used a 64-channel EEG cap with active Ag/AgCl electrodes that followed the international 10-10 system to record EEG signals. The FCz and FPz channels were set as the reference and ground electrodes, respectively. Brain Vision/Recorder software (BrainProduct GmbH, Germany) was used to collect the EEG signals, which were then operated using the MATLAB 2018a software. The researchers made sure that the impedance of all electrodes was kept below 10 $k\Omega$. The researchers randomly presented 22 blocks of 12 words and a rest class. Each of the 22 participants contributed 1,300 samples, consisting of 100 samples per category. The study was approved by the Institutional Review Board of the Korea University [KUIRB-2019-0143-01] and was conducted in accordance with the Declaration of Helsinki.

\subsubsection{Preprocessing}
This research used a variety of preprocessing techniques to ensure the accuracy of the EEG data. Initially, a bandpass filter was used to filter signals between 0.5 and 125 Hz, with additional notch filtering at 60 and 120 Hz to eliminate power line interference. Subsequently, a common average reference method was used to reference the data and reduce any noise present. To remove ocular and muscular artifacts caused by movement or sounds, automatic electrooculography and electromyography removal methods were employed. After the artifacts were removed, the EEG signals were chosen in the high-gamma frequency band to train the model and analyze the data. The data set was then epoched into 2-second segments, with a baseline correction applied 500 ms before the task. All preprocessing steps were performed using MATLAB-based tools, such as the OpenBMI Toolbox \cite{leeMH2019eeg,lee2021decoding} or BBCI Toolbox \cite{krepki2007berlin}.

\section{RESULTS AND DISCUSSION}
In this study, we compared the performance of our proposed method with three established approaches: DeepConvNet \cite{schirrmeister2017deep}, EEGNet \cite{lawhern2018eegnet}, and the method introduced by Lee et al. \cite{lee2020neural}, in the context of the decoding of the EEG signal from spoken speech. The results, presented in Table 1, show that our method outperformed the other three in terms of both accuracy and area under the curve (AUC). The average accuracy of our approach was 72.33 \%, with a standard deviation of 7.51 \%, while the average AUC was 93.22 \%, with a standard deviation of 3.18 \%. These figures are significantly higher than those of the compared methods. Specifically, the approach of DeepConvNet, EEGNet and Lee et al. yielded average average accuracies of 32.34 \%, 42.73 \%, and 57.06 \%, and average AUCs of 73.00 \%, 81.00 \%, and 83.01 \%, respectively. This indicates the superior ability of our proposed method in decoding EEG signals related to spoken speech.

Our research yielded unexpected results, especially since the more traditional approach of Lee et al. \cite{lee2020neural}, which combines a common spatial pattern with the support vector machine, outperformed popular EEG decoding methods such as EEGNet \cite{lawhern2018eegnet} and DeepConvNet \cite{schirrmeister2017deep}. These methods have been extensively used for motor imagery and event-related potentials. Our findings emphasize the importance of selecting a suitable model architecture that is compatible with the task, the EEG paradigms used, and other relevant factors.


\begin{table}[t]
\centering
\renewcommand{\tabcolsep}{3mm}
\caption{Accuracy and AUC scores for imagined speech classification}
\label{tab:result}
{
\begin{tabular}{lcc}
\hline
\textbf{Subject}     & \textbf{Accuracy (\%)} & \textbf{AUC (\%)}\\ 
\hline
\textbf{DeepConvNet} & 32.34 $\pm$ 5.10 & 73.00 $\pm$ 4.00 \\
\textbf{EEGNet} & 42.73 $\pm$ 3.80 & 81.00 $\pm$ 4.19 \\
\textbf{Lee et al.} & 57.06 $\pm$ 6.52 & 83.01 $\pm$ 5.10 \\
\textbf{Diff-E} & \textbf{72.33 $\pm$ 7.51} & \textbf{93.22 $\pm$ 3.18} \\
\hline
\end{tabular}
}
\end{table}

\begin{table}[t]
\centering
\caption{Efficacy of Each Component in Diff-E: An Ablation Study Assessing the Individual Contributions of DDPM and CAE}

{
\begin{tabular}{lcc}
\hline
\textbf{Components}     & \textbf{Accuracy (\%)}  & \textbf{AUC (\%)}\\ 
\hline
\textbf{Diff-E} & \textbf{72.33 $\pm$ 7.51} & \textbf{93.22 $\pm$ 3.18} \\
\textbf{w/o DDPM} & 52.11 $\pm$ 8.98 & 90.19 $\pm$ 5.11 \\
\textbf{w/o DDPM \& $\mathcal{D}_\psi$} & 51.11  $\pm$ 8.80 & 66.53 $\pm$ 4.54 \\

\hline
\end{tabular}%
}
\label{tab:ablation}
\end{table}
\section{CONCLUSION}
This research marks a significant advancement in the field of EEG signal decoding, particularly with regard to the challenge of interpreting spoken language. Our study introduces an innovative application of generative models that showcases improved performance over more conventional neural network approaches such as DeepConvNet and EEGNet. The findings suggest that generative models could be instrumental in improving the processing of EEG signals and could potentially be adapted for wider applications within this scientific area.

Furthermore, our research underscores the critical role of model architecture selection in EEG decoding tasks. The observed disparities in the performance of established methods like DeepConvNet and EEGNet, when compared to our generative model approach, underscore this point. The choice of model must be informed by a thorough understanding of the characteristics of the EEG data and the specific requirements of the decoding task.

In summary, our research offers a promising avenue for the accurate interpretation of EEG signals related to spoken language. This has far-reaching implications for the progression of brain-computer interfaces, potentially enhancing communication capabilities and assistive technologies. Additionally, it furthers the knowledge and application of deep learning in EEG analysis, potentially setting a precedent for future research in this vital area.
\bibliographystyle{jabbrv_IEEEtran}
\bibliography{REFERENCE}

\end{document}